\documentclass[atmp,reqno]{ipart_v1}

\makeatletter
\def\lastpage@putlabel{}
\makeatother
\usepackage{lastpage,hyperref}

\usepackage{comment}
\usepackage[svgnames]{xcolor}
\usepackage{graphicx} 
\usepackage{amssymb}
\usepackage{amsmath}
\usepackage{placeins}
\usepackage{tikz}
\usetikzlibrary{positioning}
\usetikzlibrary{calc}
\usetikzlibrary{fit}
\usepackage[framemethod=tikz]{mdframed}
\usepackage{wrapfig}
\usepackage{subcaption}

\newcommand{\set}[1]{\left\{ #1 \right\}}

\newcommand{\R}{\mathbb{R}} 

\newcommand{\Z}{\mathbb{Z}}
\newcommand{\N}{\mathbb{N}}

\newcommand{\paren}[1]{\left( #1 \right)}
\newcommand{\squaren}[1]{\left[#1 \right]}

\newcommand{\md}{\text{ mod }}

\newcommand{\sube}{\subseteq}

\newcommand{\bm}[1]{\begin{bmatrix}#1\end{bmatrix}}
\newcommand{\ve}{\mathbf}
\newcommand{\relu}{\operatorname{ReLU}}

\newcommand{\newpara}{\newline\newline}

\newcommand{\X}{\mathcal{X}}
\newcommand{\floor}[1]{\left\lfloor #1 \right\rfloor}

\newcommand{\prt}[2]{\squaren{\begin{array}{#1} #2\end{array}}}

\newcommand{\startperm}{\Delta}
\newcommand{\layernum}{N_L}  

\newmdenv[innerlinewidth=0.5pt, roundcorner=4pt,linecolor=red,innerleftmargin=6pt,
innerrightmargin=6pt,innertopmargin=6pt,innerbottommargin=6pt]{irrelevant}
\newmdenv[innerlinewidth=0.5pt, roundcorner=4pt,linecolor=orange,innerleftmargin=6pt,
innerrightmargin=6pt,innertopmargin=6pt,innerbottommargin=6pt]{old}
\newmdenv[innerlinewidth=0.5pt, roundcorner=4pt,linecolor=Goldenrod,innerleftmargin=6pt,
innerrightmargin=6pt,innertopmargin=6pt,innerbottommargin=6pt]{didntwork}
\newmdenv[innerlinewidth=0.5pt, roundcorner=4pt,linecolor=Green,innerleftmargin=6pt,
innerrightmargin=6pt,innertopmargin=6pt,innerbottommargin=6pt]{current}

\title{Learning the symmetric group: large from small}
\author{Max Petschack, Alexandr Garbali and Jan de Gier}
\date{\today}

\begin{document}

\begin{abstract}
Machine learning explorations can make significant inroads into solving difficult problems in pure mathematics. One advantage of this approach is that mathematical datasets do not suffer from noise, but a challenge is the amount of data required to train these models and that this data can be computationally expensive to generate. Key challenges further comprise difficulty in a posteriori interpretation of statistical models and the implementation of deep and abstract mathematical problems. 

We propose a method for scalable tasks, by which models trained on simpler versions of a task can then generalize to the full task. Specifically, we demonstrate that a transformer neural-network trained on predicting permutations from words formed by general transpositions in the symmetric group $S_{10}$ can generalize to the symmetric group $S_{25}$ with near 100\% accuracy. We also show that $S_{10}$ generalizes to $S_{16}$ with similar performance if we only use adjacent transpositions. We employ identity augmentation as a key tool to manage variable word lengths, and partitioned windows for training on adjacent transpositions. Finally we compare variations of the method used and discuss potential challenges with extending the method to other tasks.
\end{abstract}

\maketitle

\section{Introduction}

Transformer-based AI for mathematics is a fast-developing field. The effectiveness of transformers have been tested on variety of mathematical problems, including arithmetic tasks \cite{NogueiraJL2021}, linear algebra \cite{Charton2021}, knot theory \cite{GukovHRS:2020}, and pattern recognitions \cite{ChartonEWW}. 

Inspired by the works \cite{NandaCLSS2023,ChughtaiCN2021} our goal is to further investigate whether transformers can efficiently learn group theory, and whether features learned from relatively small training sets are in some sense universal, so that large and more complicated groups can be studied by scaling up.  

Motivated by machine learning experiments on the challenging unknot problem \cite{GukovHRS:2020}, we focus on the simpler problem in the context of the symmetric group $S_n$ with the goal of training a transformer to take as input a word from $S_n$ expressed in terms of transpositions, and return a prediction for the corresponding permutation of $(1,\ldots,n)$ without hardcoding the action of transpositions.  Generalizations and future work may provide insights in the word problem for finitely generated groups (which is of course solved for the symmetric group, see e.g. \cite{Garsia2002}) and applications to NP-hard computational  problems related to exchanges of two consecutive sequences of genes in a genome in comparative genomics \cite{BafnaP1998,BulteauFR}.

Our aim here thus is to investigate whether $S_n$ can be learned by training on smaller structures such as subgroups $H < S_n$ with $H\simeq S_k$ and $k \ll n$. In this experiment specifically, we develop a method for training a transformer on $S_{10}$ and then have it be capable of predicting permutations from words in the higher order groups $S_{16}$ and $S_{25}$.

\subsection{Related work on symmetric group}
Under continued training we assume that the network generalizes, i.e. gradually shifts from memorization to learning an underlying algorithm. While we achieve test accuracy of nearly 100\%, at the time of writing it is an open question whether our models have discovered a general algorithm or learned specific representations.

Interpretability of group multiplication in small permutation groups has been studied in a series of works \cite{ChughtaiCN2021,StanderYFB2023,Wu2024} (see also \cite{NandaCLSS2023}). Analysis of symmetric group models using mechanistic interpretability can be done using different circuit level approaches \cite{Olah2020,Elhage2021,Olsson2022,Quirke2024,Zhang2023,Zhong2023,Gross2024}. Examining activations of individual neurons and intervening in various parts of the network can help us to discover circuits that implement the algorithm. We aim to perform a similar analysis of our models in a future work.

In contrast with previous works on learning tasks related to the symmetric group, we investigate whether a model is able to generalize from training on $S_m$ data to testing on $S_n$ data with $m< n$. This is out of distribution (OOD) learning which in the context of mathematics learning tasks has also been studied in \cite{Charton2022}. We test OOD learning in the context of the symmetric group, using a scalable presentation with general transpositions as group generators, as well as a less scalable, more local approach using only elementary transpositions. One motivation for this study is to identify an effective machine learning approach to improve on the unknot problem \cite{GukovHRS:2020}.

The code used for this project is available on GitHub at \cite{Petschack_github}.

\section{Permutations and the symmetric group}

\subsubsection*{Permutation} A permutation is a rearrangement of an ordered set. Here we will consider permutations $\sigma$ of the set $S=\{1,\ldots,n\}$, and use one-line notation $(\sigma(1),\ldots,\sigma(n))$ for the permutation $\sigma: S\to S$ in which each element $i$ is replaced by the corresponding $\sigma(i)$. We denote by $\Pi_n$ the set of all permutations of a set with $n$ elements. 

\subsubsection*{Symmetric group} The set $\Pi_n$ can be endowed with a multiplication to form the symmetric group $S_{n}$, where the group operation for two permutations $\sigma$ and $\tau$ in the group $S_n$ is the product $\pi = \sigma \tau$ defined by,
\begin{equation}
\pi (i)=\sigma (\tau (i))
\end{equation}
For example, for $\sigma=(2,1,3)$ and $\tau=(3,2,1)$ we find $\pi=\sigma\tau=(3,1,2)$ and $\pi'=\tau\sigma=(2,3,1)$.

\subsubsection*{Transposition} A transposition $s_{i,j}$ is a permutation in which elements $i$ and $j$ are interchanged when multiplied on the right,
\begin{equation}
\label{eq:transp}
(\ldots,\sigma(i),\ldots,\sigma(j),\ldots) \cdot s_{i,j} = (\ldots,\sigma(j),\ldots,\sigma(i),\ldots). 
\end{equation}

\subsubsection*{Adjacent transposition} The transposition $s_{i}:=s_{i,i+1}$ that exchanges adjacent elements in a permutation is called an adjacent transposition.  
\bigskip

Every group element $w\in S_n$ can be written as a word in terms of products of transpositions,
\begin{equation}
\label{eq:factor}
w=s_{i_1,j_1} s_{i_2,j_2}\cdots s_{i_\ell,j_\ell},
\end{equation}
where $\ell$ is the length of the word. A word $w$ acts on a permutation by acting from the right as in \eqref{eq:transp}. This expression is not unique, different factorisations in terms of transposition may represent the same group element as a result of the relations
\begin{align}
s_{i,j} &= s_{i}s_{i+1}\cdots s_{j-2} s_{j-1} s_{j-2} \cdots s_{i},\nonumber \\
s_is_{i+1}s_i &= s_{i+1}s_is_{i+1}, \label{group-rel}\\
s_i^2 &=1.\nonumber
\end{align}
A word is called reduced if it cannot be written using a smaller number of transpositions. 

Adjacent transpositions generate the full symmetric group, and the Coxeter presentation of $S_n$ in terms of adjacent transpositions is given by
\begin{multline}
S_{n}=\left\langle s_{1},\ldots ,s_{n-1}\,|\, s_{i}s_{i+1}s_{i}=s_{i+1}s_{i}s_{i+1}\right.,\\
\left.s_{i}s_{j}=s_{j}s_{i}{\text{ for }}|i-j|\geq 2,\, s_{i}^{2}=1\right\rangle.
\end{multline}

\section{What are we learning?}
The experiment reported in this note investigates whether a transformer can learn the symmetric group $S_n$ by training only on subgroups isomorphic to $S_m$ with $m< n$. That is, given a word $w\in S_n$ can the corresponding permutation $(\sigma(1),\ldots,\sigma(n)):=(1,\ldots,n)\cdot w$ be correctly predicted by a transformer without explicitly hardcoding the group relations \eqref{group-rel} and training only on permutations that do not permute more than $m<n$ elements.

We take two approaches, one in which input words $w$ are factorised using general transpositions as in \eqref{eq:factor}, and one in which only factorisations in adjacent transpositions are allowed. We stress that these two approaches should be viewed as different problems. In the symmetric group literature it is known that decompositions of words into general transpositions and into adjacent transpositions are associated to algorithms of different complexity (see for example \cite{BayerDiaconis1992,Lacoin2016}). 

\subsubsection*{Variable word length and identity augmentation} The reduced word length $\ell$ in the factorisation \eqref{eq:factor} is variable for different elements. It ranges between $\ell=0$ for the identity element and $\ell_\mathrm{max}=n-1$ for the longest permutation element. If we consider factorisations of words into adjacent transpositions $s_i$ then the maximal word length corresponding to the longest permutation is $\ell_\mathrm{max}=n(n-1)/2$. In order to deal with variable word length we fix $N = \ell_\mathrm{max}$ and write each word in unreduced form using $N$ transpositions, i.e. reduced words of length $\ell<N$ are augmented with sufficiently many transpositions that amount to identities under group relations. The transformer will need to learn the group relations.

\subsection{General transpositions}
\subsubsection*{Tokenization} We tokenize a word in the symmetric group as an integer tuple corresponding to a factorisation into general transpositions. Let $N\in\N$ be the maximum word length and $n$ the maximum group size. We take the input $\ve x = (x_1,\ldots, x_N)$ to the model as a vector of integers 
\begin{equation}
\label{eq:Xdef}
\ve x\sube\X^N,\qquad \X=\set{0,\dots,n^2-1}, 
\end{equation}
corresponding to a word in $S_n$ via the map $w:\X^{N}\to\ S_n$, where
\begin{equation}
\begin{split}
    w(x_1,\dots,x_N) &= s_{i_1,j_1}\cdots s_{i_N,j_N},\\
    x_k &= i_k-1 + n(j_k-1).   
\end{split}
\label{eq:input_general}
\end{equation}
so that
\begin{equation}
i_k = 1+\floor{x_k/n},\qquad j_k = 1+ (x_k\bmod n).
\end{equation}
    
It is worth noting that there are other possible ways that $w$ could be tokenized, but we observed that the one here works best in our setup. 

\subsubsection*{Training on small subgroups}
We trained only on words that permute at most $m<n$ elements and tested on the map $\Phi_N : \X^N \to \Pi_n$ given by
\begin{equation*}
\hspace{-6cm}\Phi: \ve x \mapsto \ve p,
\end{equation*}
where $\ve p =(p_1,\ldots, p_n)$ is the permutation obtained by applying $w(\ve x)$ to $(1,\ldots,n)$. 
In our experiment $n=25$ and $m=10$.

In order to train using only information from smaller subgroups, we construct words in $S_n$ that permute at most $m$ elements and represent these as factorised expressions into general transpositions that are element of $S_n$. This is implemented by first generating a word in $S_m$ represented by a tuple $(\ve i, \ve j) = (i_1,j_1,\ldots, i_{n-1},j_{n-1}) \in \{1,\ldots,m\}^{2(n-1)}$ according to its factorisation.\footnote{For implementation purposes we generate here unreduced words of length $n-1$ which is larger than they need to be for $S_m$.} We then convert $(\ve i, \ve j)$ to a tuple corresponding to a word in the larger group $S_n$ by \textbf{relabeling} using the map
\begin{equation}
    \ve (i_1,j_1,\dots,i_{n-1},j_{n-1})\mapsto(\sigma(i_1),\sigma(j_1)\dots,\sigma(i_{n-1}),\sigma(j_{n-1})),
    \label{eq:relabel}
\end{equation}
with $\sigma\in S_n$ a random permutation of $\{1,\ldots,n\}$. The output in \eqref{eq:relabel} is then mapped into input form $\ve x$ using the inverse of \eqref{eq:input_general}.

\subsection{Adjacent transpositions}
\subsubsection*{Tokenization}
In the case of words that are factorised using only adjacent transpositions, 
\begin{equation}
w=s_{i_1} \cdots s_{i_N},\qquad N=n(n-1)/2,
\end{equation}
with $i_k\in \{0,\ldots, n-1\}$ and $s_0 \equiv 1$. We take as input simply $\ve x = (x_1,\ldots, x_N)$ with $x_k = i_k$. In this experiment we take $n = 16$.

\subsubsection*{Training on small subgroups}
A naive attempt to implement training on smaller subgroups $S_m$ is the \textbf{window method} which is a modified version of relabeling compatible with elementary transpositions, requiring that the possible transpositions that we choose must be adjacent, i.e we relabel $\{s_1,\ldots,s_{m-1}\}$ to $\{s_{k+1},\ldots,s_{m+k-1}\}$ for some choice of $k$. This naive version of the window method does not perform well in practice when testing on larger groups. The intuition behind this observation is that the window technique fails because the transformer is lazy; instead of learning the larger group, it learns the smaller group and just figure out where the window is. 

In order to combat this, we need to make it harder for the transformer to work out where the window is. We do this using a \textbf{partitioned window method} that accommodates for multiple windows of different lengths. When generating a word we would first pick a composition of $m$ at random with the restriction that the smallest part is three to ensure inclusion of nontrivial group relations. This composition then provides a multi-window configuration. For example, if $m=12$ and we pick the composition $12=3+6+3$, then this would give us three windows, one of length six and two of length three. We then pick admissible offsets for each of these windows so that they fit into an interval of length $n$, allowing them to potentially overlap. Finally, we generate a word using relabeling to elementary transpositions within these windows.

This principle was implemented in a slightly different way for compatibility reasons. First, we generate a word in $S_m$ of length $N=n(n-1)/2$. Let $\ve x=(x_1,\dots,x_N)$ represent the input form of this word. Then, we generate a composition $\mu =(\mu_1,\dots,\mu_k)$ where $\mu_1+\dots+\mu_k=m$, representing the possible window sizes, and choose a corresponding set of admissible offsets $O=\{o_1,\ldots,o_k\}$. Finally, we generate a set of integers $\{\ell_1,\ldots,\ell_N\}$ uniformly at random with $\ell_i \in \{1,\ldots,k\}$, and perform the map
\begin{equation*}
    \ve x\mapsto\paren{o_{\ell_1}+(x_1\md \mu_{\ell_1}),\dots,\,o_{\ell_N}+(x_N\md \mu_{\ell_N})}.
\end{equation*}
In this experiment we take $n = 16$ and $m=10$.

\begin{figure}[h]
    \centering
    \begin{tikzpicture}[
        tn/.style={
            rectangle, 
            draw=black, 
            scale=0.7,
            align=center,
            below=5mm of input
        },
        box/.style = {
            draw,
            black,
            inner sep=10pt,
            rounded corners=5pt,
            minimum width=40mm
        },
    ]
        \pgfdeclarelayer{background}
        \pgfdeclarelayer{foreground}
        \pgfsetlayers{background,main,foreground}

        \begin{pgfonlayer}{foreground}
        \node (input) {Input};
        \node (tembedding) [tn, below=5mm of input, fill=MistyRose] {Token\\embedding};
        \draw[->, shorten >=3pt] (input) -- (tembedding);
        
        \node (plus) [circle, below=5mm of tembedding, draw=black, inner sep=0.5mm] {$+$};
        \draw[->, shorten <=2pt, shorten >=2pt] (tembedding) -- (plus);

        \node (pembedding) [tn, left=1cm of plus, fill=Azure] {Positional\\encoding};
        \draw[->, shorten <=2pt, shorten >=2pt] (pembedding) -- (plus);





        \node (branch1) [circle, below=1cm of plus, draw=white, inner sep=0.5mm] {};

        \node (attention) [tn, below=5mm of branch1, fill=Bisque] {Multi-head attention};
        \draw[->, shorten <=2pt, shorten >=3pt] (plus) -- (attention);

        \node (addnorm1) [tn, below=5mm of attention,
        fill=Cornsilk] {Add+norm};
        \draw[->, shorten <=2pt, shorten >=3pt] (attention) -- (addnorm1);

        \coordinate[left=15mm of branch1]  (branch1left) ;
        \coordinate[left=8.02mm of addnorm1]  (addnorm1left) ;
        \draw[->, shorten <=-2pt, shorten >=3pt=90] (branch1) -- (branch1left) -- (addnorm1left) -- (addnorm1);

        \node (branch2) [circle, below=5mm of addnorm1, draw=white, inner sep=0.5mm] {};

        \node (ffwd) [tn, below=5mm of branch2, fill=PaleTurquoise] {Feed forward};
        \draw[->, shorten <=2pt, shorten >=3pt] (addnorm1) -- (ffwd);

        \node (addnorm2) [tn, below=5mm of ffwd, fill=Cornsilk] {Add+norm};
        \draw[->, shorten <=2pt, shorten >=3pt] (ffwd) -- (addnorm2);

        \coordinate[left=15mm of branch2]  (branch2left) ;
        \coordinate[left=8.02mm of addnorm2]  (addnorm2left) ;
        \draw[->, shorten <=-2pt, shorten >=3pt=90] (branch2) -- (branch2left) -- (addnorm2left) -- (addnorm2);
        \end{pgfonlayer}{foreground}

        \begin{pgfonlayer}{background}
        \node[box,fit=(branch1)(addnorm2)] {};
        \end{pgfonlayer}{background}

        \begin{pgfonlayer}{foreground}

        \node (repeat) [right=2cm of branch2] {$\times\layernum$};

        \node (linear) [tn, below=8mm of addnorm2, fill=Lavender] {Linear};
        \draw[->, shorten <=2pt, shorten >=3pt] (addnorm2) -- (linear);

        \node (softmax) [tn, below=5mm of linear, fill=Honeydew] {Softmax};
        \draw[->, shorten <=2pt, shorten >=3pt] (linear) -- (softmax);

        \node (output) [below=5mm of softmax] {Output probabilities};
        \draw[->, shorten <=2pt, shorten >=-1pt] (softmax) -- (output);
        
        \end{pgfonlayer}{foreground}
        
    \end{tikzpicture}
    
    \caption{Transformer architecture used in this project}
    \label{fig:transformer}
    
\end{figure}
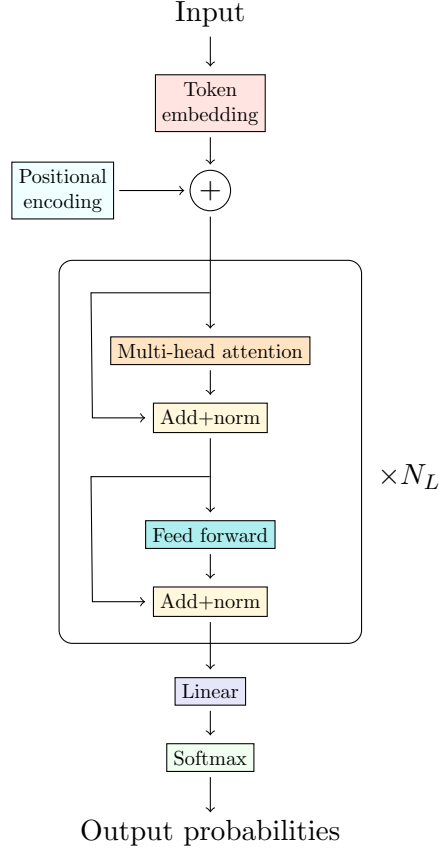

\section{Model architecture}

\subsection{The input step}
Given $\ve x$ as in \eqref{eq:Xdef}, we construct the input to the transformer (called the context window). It is comprised of $\ve x$, plus additional tokens that are used to track the transformer's progress so far. Let $C\in \N$ denote the length of the context window. The number $C$ depends on the choice of presentation of $w$ --- in the case were we use general transpositions $C=N+n$ when the full permutation has been predicted. The context window is then given by

\begin{equation*}
M^{(I)}(\ve x, \ve p)= (x_1,\dots,x_N,p_1,\dots,p_{k}).
\end{equation*}

Here $\ve p\in\Z^ k$ with $k\le n$ denotes the part of the permutation that the model has predicted so far. The model is autoregressive; it predicts a single token of the permutation at a time. This is then fed back into the model (ie. added to $\ve p$) until all tokens have been predicted and $k=n$. We initialize $\ve p$ as $\ve p_0=(\Delta)$, where $\startperm$ is a special separator token between the word and the predicted permutation.

\subsection{The embedding step}
\label{embedding_step_explanation}

Each token has a vector embedding in some high dimensional space, which is learned by the model during training. The dimension of this space is a hyperparameter of the model, and we will denote it by $D$. Let $T\subset \Z$ denote the set of all tokens, and let $E_T:T\to\R^D$ be the mapping between the tokens and their embeddings\footnote{In practice, $E_T$ is implemented as a learned $|T|\times D$ lookup table, where the embedding of the $i$th token is given by the $i$th row of the table.}. We then collect all the token embeddings into one $C\times D$ matrix via the mapping
\begin{equation*}
    M^{(E)}_T(t_1,\dots,t_C)=\bm{E_T(t_1)\\ \vdots \\ E_T(t_C)}
\end{equation*}
There is also a second embedding associated with the position of the token, which is a learned $C\times D$ matrix which we will denote with $E_P$. The position embedding, which does not depend on the input, gets added to the matrix found from the token embeddings. The full embedding step can therefore be written as a single mapping, $M^{(E)}:\Z^C\to M_{C\times D}(\R)$, where
\begin{equation*}
    M^{(E)}(t_1,\dots,t_C)=M^{(E)}_T(t_1,\dots,t_C)+E_P.
\end{equation*}


\section{Results} 

\begin{figure}[h]
\centering
\begin{subfigure}{0.4\textwidth}
    \includegraphics[width=\textwidth]{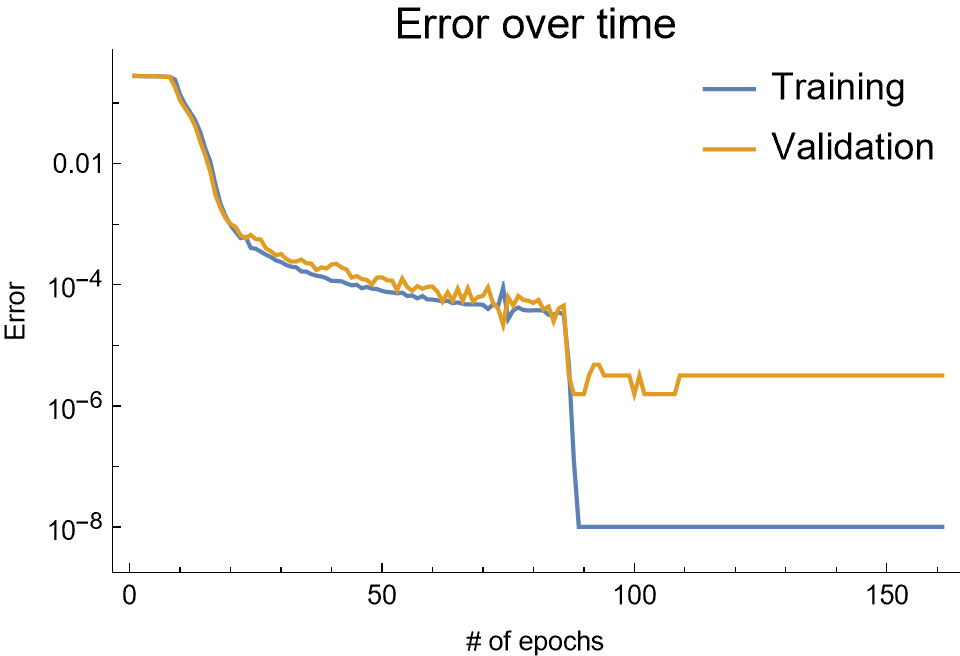}
    \caption{Training (blue) and validation (orange) error (log scale).}
    \label{fig:general_error}
\end{subfigure}\qquad
\begin{subfigure}{0.4\textwidth}
    \includegraphics[width=\textwidth]{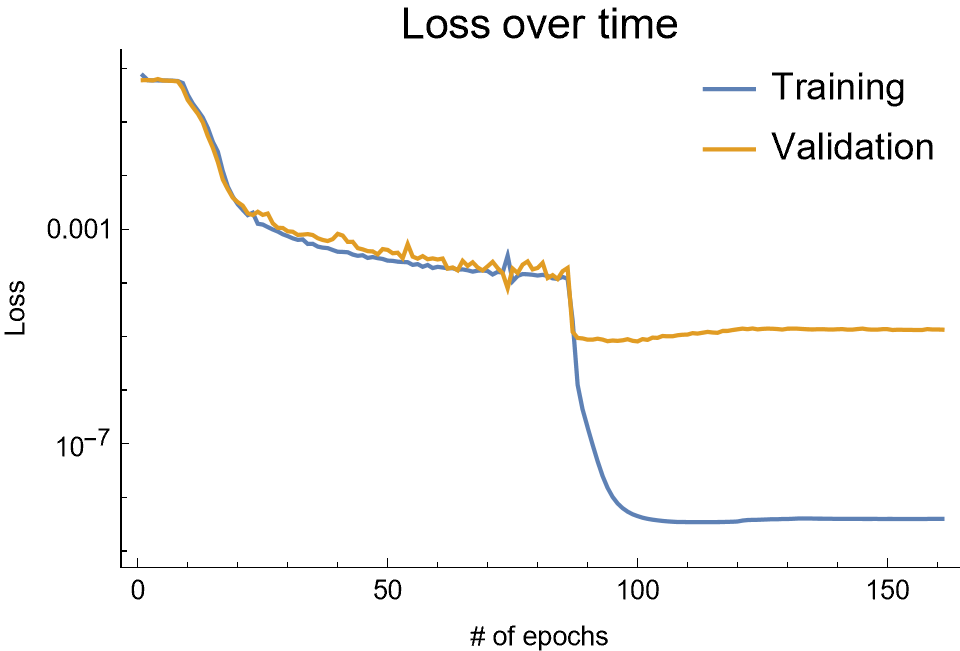}
    \caption{Training (blue) and validation (orange) loss (log scale).}
    \label{fig:general_loss}
\end{subfigure}
        
\caption{Error and loss for training and validation using general transpositions.}
\label{fig:general_performance}
\end{figure}

\begin{figure}[h]
\centering
\begin{subfigure}{0.4\textwidth}
    \includegraphics[width=\textwidth]{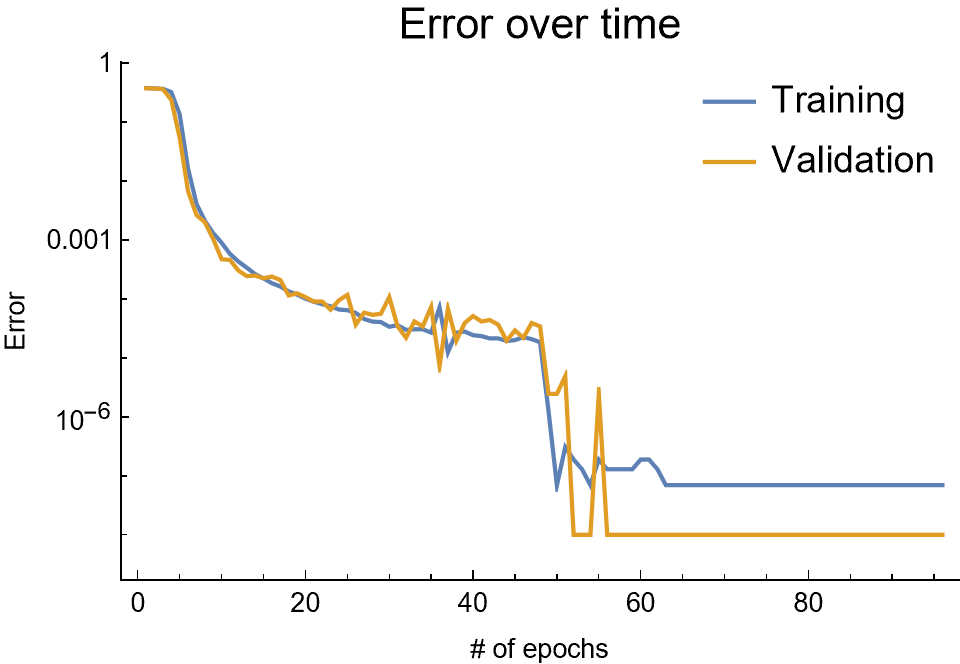}
    \caption{Training (blue) and validation (orange) error (log scale).}
    \label{fig:adjacent_error}
\end{subfigure}\qquad
\begin{subfigure}{0.4\textwidth}
    \includegraphics[width=\textwidth]{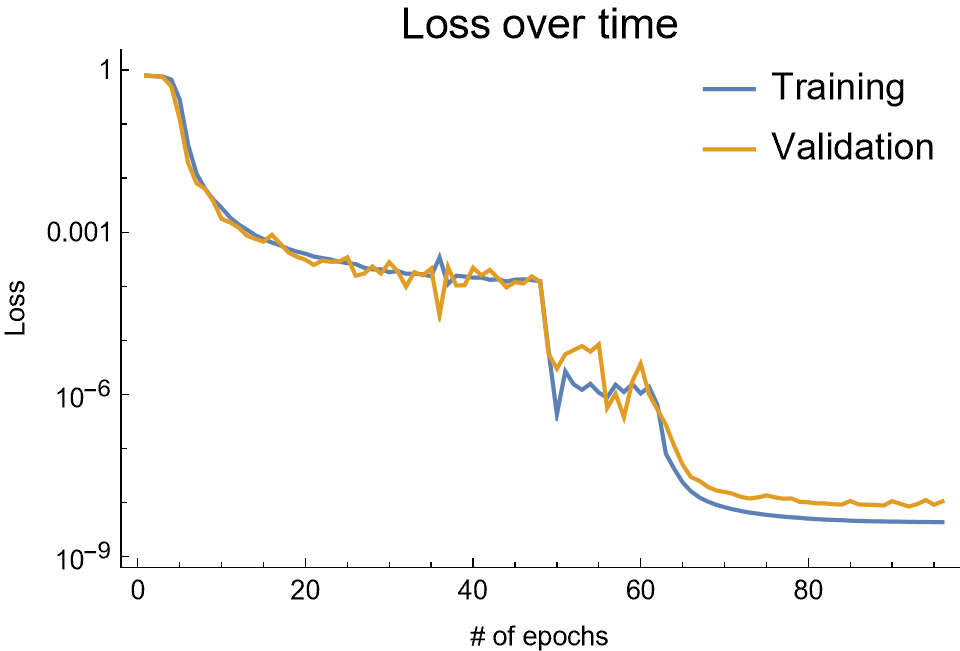}
    \caption{Training (blue) and validation (orange) loss (log scale).}
    \label{fig:adjacent_loss}
\end{subfigure}
        
\caption{Single token error and loss for training and validation using adjacent transpositions.}
\label{fig:adjacent_performance}
\end{figure}

\subsection{Performance}
To test performance of our model we measure cross-entropy loss as well as the error for each epoch where
\begin{equation}
\text{error} = \frac{\# \text{incorrect predictions}}{\# \text{test datapoints}}.
\end{equation}  

For general transpositions we train on $S_{10}$ and test on $S_{25}$. We obtained zero training loss and near perfect test performance as shown in Figure~\ref{fig:general_performance}. For adjacent transpositions, where we expect the complexity to be harder \cite{BayerDiaconis1992,Lacoin2016}), we tested on $S_{16}$ and obtained similar performance, see Figure~\ref{fig:adjacent_performance}. We elaborate on validation and test data in Section~\ref{ap:datasets}.

\subsection{Interpretation}

\begin{figure}[h]
    \centering
    \begin{subfigure}[][200pt][t]{0.4\textwidth}
        \includegraphics[width=\textwidth]{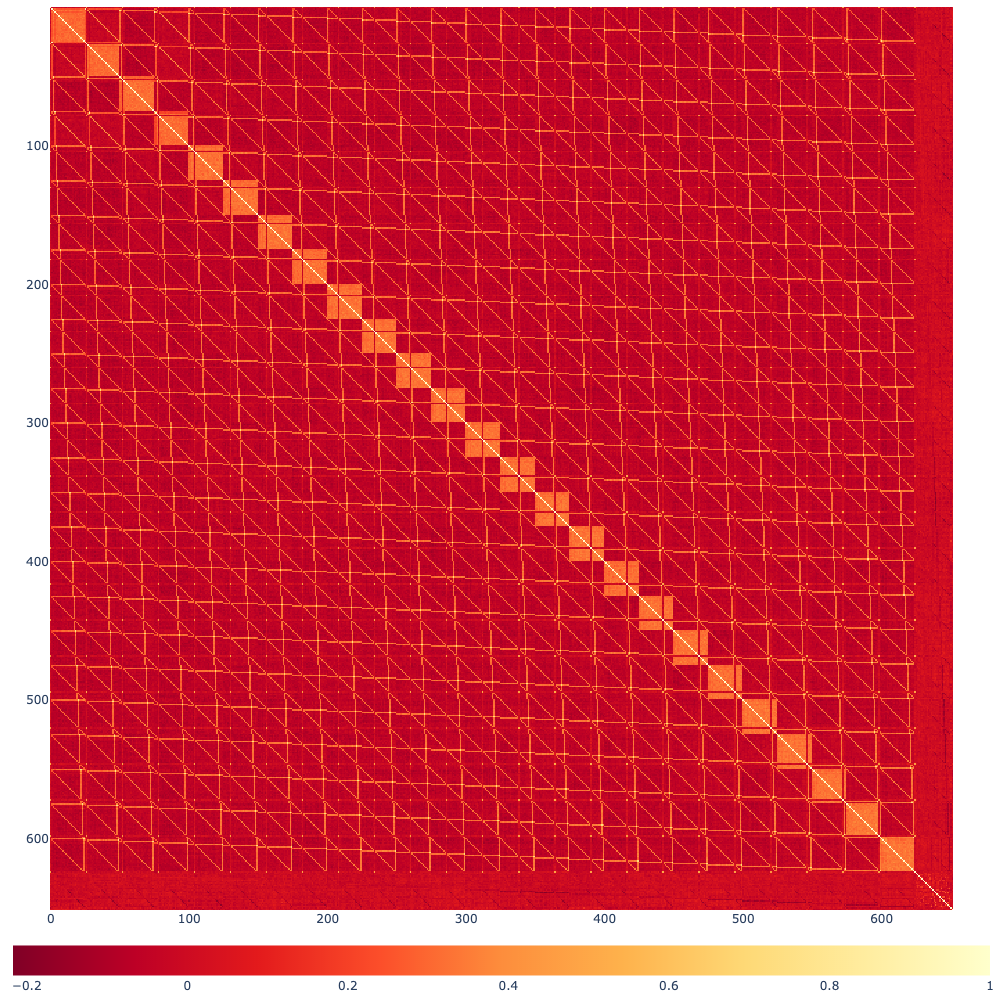}
        \caption{Token embedding for $n=25$. The first $n^2$ rows and columns correspond to transposition tokens, the last $n$ rows and columns to permutation tokens.}
        \label{fig:general_heatmaps_tok}
    \end{subfigure}
    \qquad
    \begin{subfigure}[][200pt][t]{0.4\textwidth}
        \includegraphics[width=\textwidth]{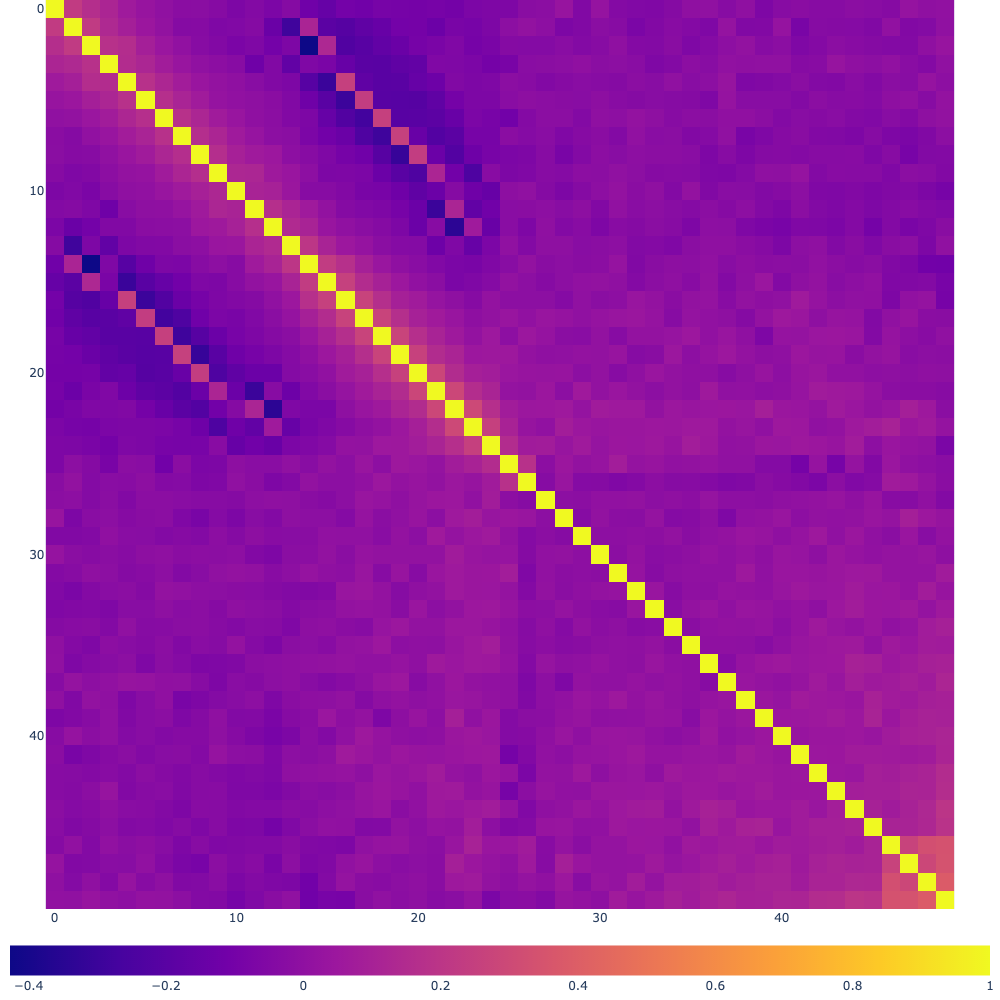}
        \caption{Position embedding for $n=25$. The first $n-1$ rows and columns correspond to positions of transposition tokens and the last $n$ rows and colums to positions of permutation tokens.}
        \label{fig:general_heatmaps_pos}
    \end{subfigure}

    \bigskip
    
    \caption{Embedding self-similarity heatmaps for the case of general transpositions.}
    \label{fig:general_heatmaps}
\end{figure}

\begin{figure}[h]
    \centering
    \begin{subfigure}[][200pt][t]{0.4\textwidth}
        \includegraphics[width=\textwidth]{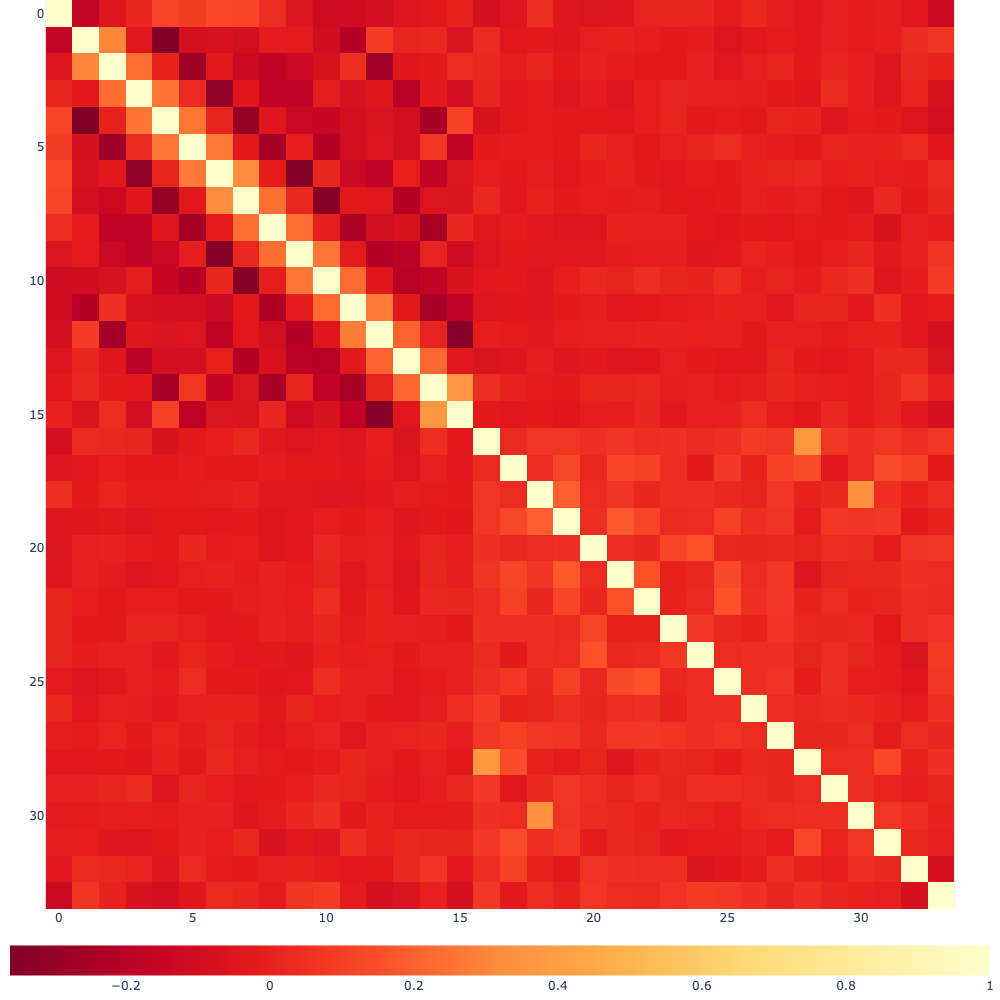}
        \caption{Token embedding for $n=16$. The first $n-1$ rows and columns correspond to transposition tokens, the last $n$ rows and columns to permutation tokens.}
        \label{fig:adjacent_heatmaps_tok}
    \end{subfigure}
    \qquad
    \begin{subfigure}[][200pt][t]{0.4\textwidth}
        \includegraphics[width=\textwidth]{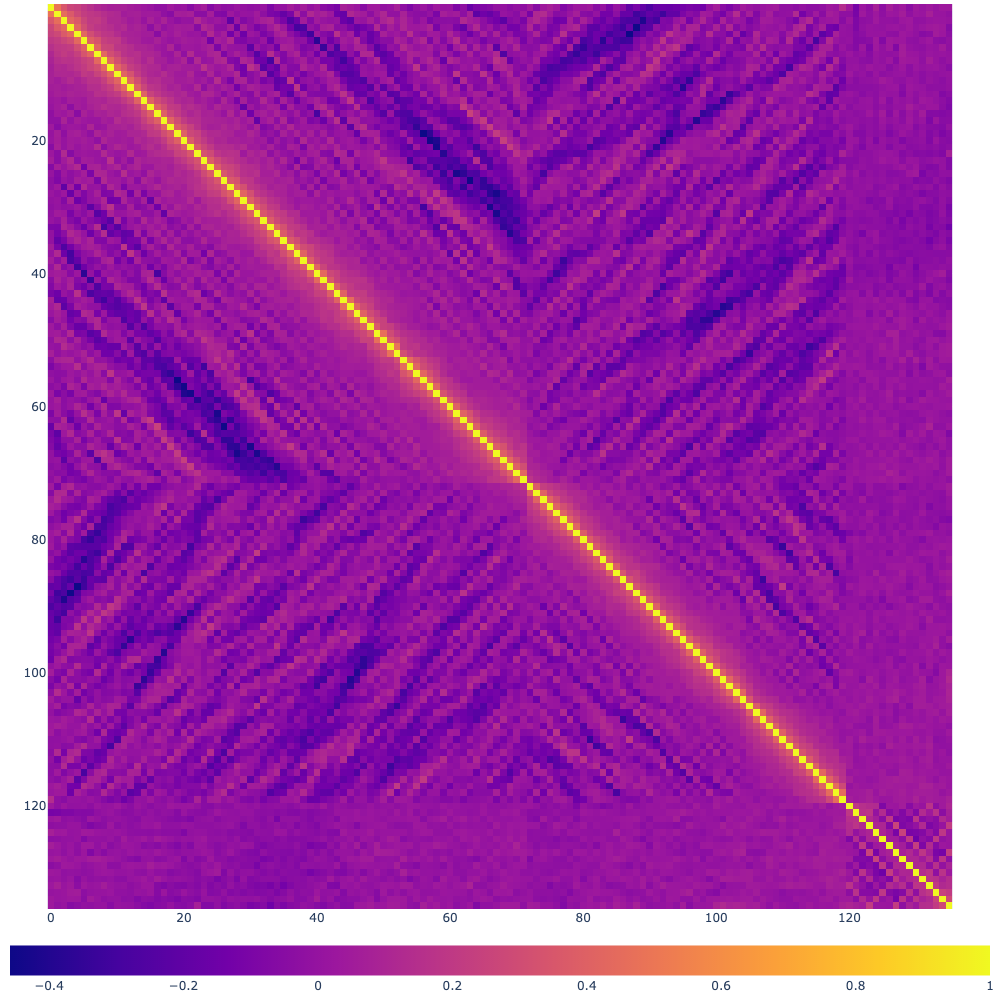}
        \caption{Position embedding for $n=16$. The first $n(n-1)/2$ rows and columns correspond to positions of transposition tokens and the last $n$ rows and colums to positions of permutation tokens.}
        \label{fig:adjacent_heatmaps_pos}
    \end{subfigure}

    \bigskip
    \caption{Embedding self-similarity heatmaps for the case of adjacent transpositions.}
    \label{fig:adjacent_heatmaps}
\end{figure}

Figures~\ref{fig:general_heatmaps} and \ref{fig:adjacent_heatmaps} depict heatmaps of the self-similarity matrices for the token embedding $M_T^{(E)}$ and position embedding $E_P$ defined in section \ref{embedding_step_explanation}. Let $A$ denote $E_P$ with normalized rows, then the matrix depicted in the image is given by $AA^T$.

\bigskip

\clearpage

We intend to further investigate interpretability of our model, and will simply highlight some preliminary structural observations. 

\subsubsection*{General transpositions} The regular lattice features in Fig.~\ref{fig:general_heatmaps_tok} indicate learned high level relationships between generators. The high-intensity lines indicate that the token corresponding to $s_{i,j}$ bears a strong relationship to that corresponding to $s_{i,k}$, and also that the tokens corresponding to $s_{i,j}$ and $s_{j,i}$ are equal. Furthermore, it is clear from the picture that all general transpositions have equal status, i.e. there is no intrinsic order among them.

The regular features in the top left block of Fig.~\ref{fig:general_heatmaps_pos} indicate that transpositions in a factorisation \eqref{eq:factor} align strongly to neighbouring transpositions. We have at the time of writing no good explanations for the high-intensity lines further away from the diagonal.

The bottom right block in Fig.~\ref{fig:general_heatmaps_pos} indicates that there is no positional structure among permutation tokens, i.e. permutation tokens are embedded in an unbiased fashion. The off-diagonal blocks simply indicate no relationship between transposition and permutation tokens.  

\subsubsection*{Adjacent transpositions} 
The token embedding structure for this case in Figure~\ref{fig:adjacent_heatmaps_tok} is simple. The position embedding in Figure~\ref{fig:adjacent_heatmaps_pos} is more interesting. The somewhat surprising block substructure within the first $120$ rows and columns in Figure~\ref{fig:adjacent_heatmaps_pos}, corresponding to the positions of transposition tokens in a word, was not always observed in experiments with other group sizes, and it is unclear what meaning, if any, should be attributed to it. Such a substructure was also observed in \cite{WangC}.  

\section{Discussion}
In this paper we report on experiments with training a transformer to predict permutations in $S_{25}$ from factorised words using \textsl{general} transpositions, as well as on predicting permutations in $S_{16}$ from factorised words using only \textsl{adjacent} transpositions. The reason for taking $n$ smaller in the case of adjacent transpositions is that training time scales proportional to $N^2$ where $N$ is the maximum word length. The length $N$ is proportional to $n^2$ for the adjacent case and to $n$ in the general case. We don't however expect the outcomes of our experiments to change for larger $n$ if given more training time.

In both cases we show near 100\% accuracy after learning from smaller subgroups isomorphic to at most $S_{10}$. We use identity augmentation to implement words of varying length.

It is well known that several statistical group properties increase in complexity when using only adjacent transpositions instead of general transpositions. Aside from the difference in maximum word length mentioned above, in our experiment the key difference between the two cases is exhibited mostly in the training design. In the case of adjacent transpositions we introduce the method of partitioned windows to train using only words in $S_{10}$ expressed in terms of adjacent transpositions within each window. This method allows for effective learning through local probing of larger words, akin to local probing of long sequences such as DNA strings.

The symmetric group is one of the most well behaved groups, and we should expect similar learning tasks such as word problems in more exotic groups to be more challenging. Our results in the case of adjacent transpositions are nonetheless promising for such more complex tasks where only local probing is available. We hope, for example, to make progress on the braid group, which is the key algebraic structure to the unknot problem.

\section*{Acknowledgments}
We warmly thank Persi Diaconis, Paul Kerr and Geordie Williamson for discussions and encouragement. MP was supported by a student summer research scholarship in the School of Mathematics and Statistics at The University of Melbourne. This research was supported by The University of Melbourne’s Research Computing Services and the Petascale Campus Initiative.

\appendix

\section{Appendix: Technical information}
\subsection{Model hyperparameters}
All runs were trained using AdamW and cross entropy loss. The model is a standard transformer architecture with masked multi-headed attention. A custom attention mask was used, which is discussed later in the appendix. During training, we used a reduce-on-plateau learning rate scheduler with a reduction factor of 0.1 and a patience of 10 epochs. Teacher forcing was used during training.
\newpara
We used the following hyperparameters for our runs. 

\begin{center}
    \begin{tabular}{cccc}
        \textbf{Transpositions} & \textbf{Dataset size} & \textbf{Context 
        length} & \textbf{Vocabulary size}\\
        \hline
        General & 8,000,000 & 50 & 652\\
        Elementary & 16,000,000 & 136 & 34\\
\\
        \textbf{Learning rate} & \textbf{Weight decay} & \textbf{Batch size} & \textbf{Embedding dimension}\\
        \hline
        0.0003 & 0.05 & 1024 & 402\\
        0.0003 & 0.05 & 1024 & 402\\
\\
        \textbf{Head count} & \textbf{Block count} & \textbf{FP format}&\textbf{Trainable parameters}\\
        \hline
        6 & 5 & bf16& 10,261,452\\
        6 & 5 & bf16& 9,799,152
    \end{tabular}
\end{center}
Here block count refers to $N_L$ as depicted in figure \ref{fig:transformer}. All the code can be found on our GitHub \cite{Petschack_github}\footnote{Transformer code is in the scaling-generator folder. Data generation code is in the fast-data-gen folder.}.

\subsection{Custom attention mask}
We would like all the tokens that make up the word to be able to attend to each other, but we also wanted to make sure that the model could not attend to future autoregression tokens. As such, we created the custom attention mask
\begin{equation}
    \prt{c|c}{
        \mathbf{1}_{N\times N}&\mathbf{0}_{N\times n}\\
        \hline
        \mathbf{1}_{n\times N}&L_{n\times n}
    },
\end{equation}
where $L$ denotes a lower triangular matrix.

\subsection{Architecture details}
This section elaborates on the architecture depicted in figure \ref{fig:transformer}. The feed forward block has the following structure:
\begin{gather*}
\text{input}\to \text{Affine}(D,4D)\to\relu\to\text{Affine}(4D,D)\to\text{output}.
\end{gather*}
Here $\text{Affine}(A,B)$ denotes an affine transformation from $A$ to $B$ dimensions, ie. a PyTorch linear layer. The final decoding layer (purple) is a $\text{Linear}(D, T)$ non-affine linear transformation. This is slightly inefficient as only a small subset of the tokens are valid outputs, but it does not make a large difference in practice.

\subsection{Validation and test sets.}
\label{ap:datasets}
The validation set is size $10^5$ and is drawn from the same distribution as the training data, i.e. it is restricted to permutations of $m<n$ elements. The test data has size $10^6$ and is drawn from all possible permutations. The model for general transpositions found the correct permutation with 100\% accuracy and the case of elementary transpositions achieved 99.993\% accuracy. All the datasets are generated independently and there is no specific check against shared elements, but in practice the sample space is so large that it never happened (this was checked several times).

When generating random words, it occasionally happens that the final permutation lies within the smaller subgroup by chance. To make sure this did not effect our results, we calculated an upper bound on the amount of test data this applied to by determining which fraction of the test words permuted at most $m$ elements. For the case of general transpositions this was 0.7315\%, and for the case of adjacent transpositions it was 2.243\%, a relatively small percentage of the data in both cases. 

\end{document}